\documentclass[10pt, twocolumn]{ICCAS2021}
 
\usepackage{diagbox}
\usepackage[%
  backend=bibtex      
 ,style=numeric-comp  
 ,sorting=none        
 ,sortcites=true      
 ,block=none
 ,indexing=false
 ,citereset=none
 ,isbn=true
 ,url=true
 ,doi=true            
 ,natbib=true         
]{biblatex}
\begin{document}

\title{Vision-based Autonomous Disinfection of High-touch Surfaces in Indoor Environments}

\author{Sean Roelofs${}^{1}$, Benoit Landry${}^{2}$, Myra Kurosu Jalil${}^{3}$, Adrian Martin${}^{3}$, Saisneha Koppaka${}^{3}$, Sindy K.Y. Tang${}^{3}$, and Marco Pavone${}^{2}$ }

\affils{ ${}^{1}$ Department of Computer Science, Stanford University \\
        ${}^{2}$ Department of Aeronautics and Astronautics, Stanford University \\
        ${}^{3}$ Department of Mechanical Engineering, Stanford University}


\abstract{
Autonomous systems have played an important role in response to the Covid-19 pandemic. Notably, there have been multiple attempts to leverage Unmanned Aerial Vehicles (UAVs) to disinfect surfaces. Although recent research suggests that surface transmission is less significant than airborne transmission in the spread of Covid-19, surfaces and fomites can play, and have played, critical roles in the transmission of Covid-19 and many other viruses, especially in settings such as child daycares, schools, offices, and hospitals. Employing UAVs for mass spray disinfection offers several potential advantages, including high-throughput application of disinfectant, large scale deployment, and the minimization of health risks to sanitation workers. Despite these potential benefits and preliminary usage of UAVs for disinfection, there has been little research into their design and effectiveness. In this work, we present an autonomous UAV capable of effectively disinfecting indoor surfaces. We identify relevant parameters such as disinfectant type and concentration, and application time and distance required of the UAV to disinfect high-touch surfaces such as door handles. Finally, we develop a robotic system that enables the fully autonomous disinfection of door handles in an unstructured and previously unknown environment. To our knowledge, this is the smallest untethered UAV ever built with both full autonomy and spraying capabilities, allowing it to operate in confined indoor settings, and the first autonomous UAV to specifically target high-touch surfaces on an individual basis with spray disinfectant, resulting in more efficient use of disinfectant.
}


\maketitle


\section{Introduction}
    UAVs have seen widespread use in response to the Covid-19 pandemic. They have delivered medical supplies, including personal protection equipment (PPE) \& Covid-19 tests, enforced quarantine restrictions, tracked crowd sizes, and even delivered Girl Scout cookies. One particular use of UAVs that has garnered a lot of attention is mass spray disinfection. UAVs with disinfectant spraying capabilities have been used in countries including China, the United States, India, Spain, France, Indonesia, and Nigeria \cite{china-disenfect, usa-disenfect, india-disenfect, spain-disenfect, france-disenfect, indonesia-disenfect, nigeria-disenfect}. Many of these UAVs were originally built for spraying agricultural fields but were re-purposed during the pandemic.
    
    Recently, scientific research has shown that the risk of contracting SARS-CoV-2 (the virus that causes Covid-19) through contact with fomites \cite{ethanol-coronaviruses}, defined as inanimate objects or surfaces contaminated with infectious agents, is relatively low. The United States Center for Diseases Control and Prevention (CDC) estimates that each contact with an infected surface has a less than 1 in 10,000 chance of resulting in an infection with SARS-CoV-2 \cite{cdc-surface-transmission}. Even so, recent modeling results suggest that fomites may be an important source of SARS-CoV-2 transmission, particularly in schools and child daycares, as the virus can persist on some surfaces for up to 72 hours \cite{covid-modeling}. Other human coronaviruses (e.g. CoV-229E, MERS-CoV and SARS-CoV) may also transmit via surface contact \cite{fomite-transmission-norovirus-h1n1-SARS-Lei, fomite-transmission-MERS, fomite-transmission-hcov-doorknobs}.
    
    In addition to coronaviruses, many other viruses may spread via fomite transmission. Norovirus, the leading cause of acute gastroenteritis that causes 19-21 million infections and 570-800 deaths annually in the United States \cite{illness-norovirus-hall}, is known to spread due to surface contamination \cite{fomite-transmission-norovirus-Repp, fomite-transmission-norovirus-h1n1-SARS-Lei, norovirus}. Many other pathogens, such as adenoviruses and influenza A H1N1 can also spread through contact with contaminated surfaces and objects \cite{fomite-transmission-norovirus-influenza-rhinovirus,fomite-transmission-adenovirus, fomite-transmission-norovirus-h1n1-SARS-Lei, h1n1, adenoviruses}. 
    
    Offices, restaurants, sports and music venues, and other public indoor spaces are prone to fomite transmission. Frequent surface disinfection could therefore reduce the spread of these pathogens. In a study that tested various surfaces inside a classroom for the presence of the human coronavirus CoV-229E, doorknobs and desk tops were frequently found to be contaminated \cite{fomite-transmission-hcov-doorknobs}. We therefore focus our efforts on the disinfection of high-touch surfaces, such as door handles, to reduce the spread of pathogens, including SARS-CoV-2, via fomite transmission.
    
\begin{figure}
    \centering
      \includegraphics[width=\columnwidth, trim=4cm 8cm 7cm 6cm, clip=true]{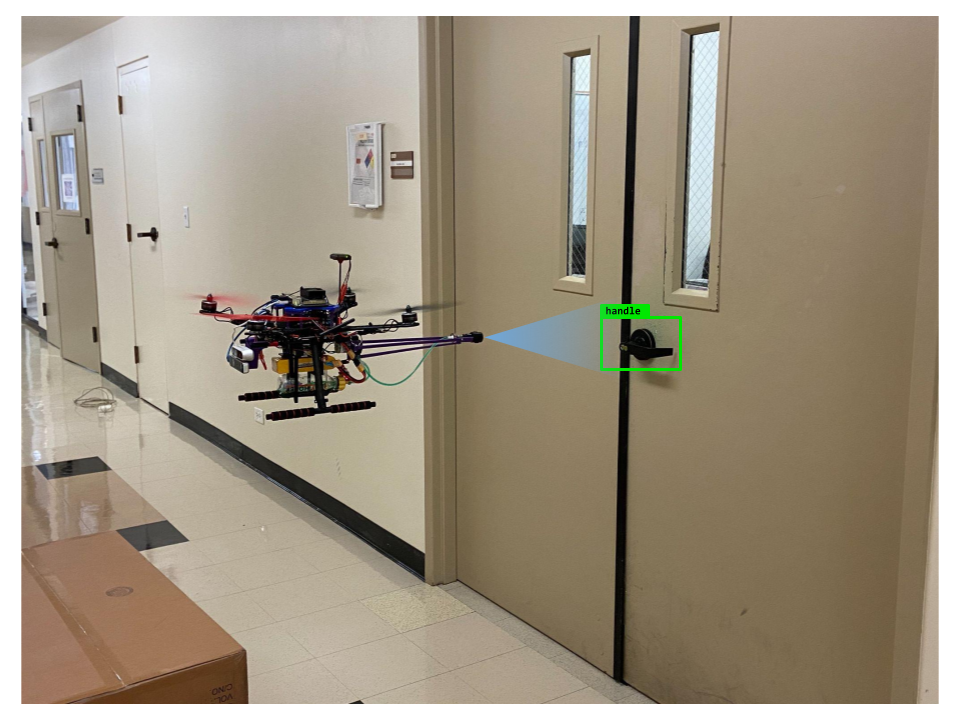}
    \caption{The UAV is capable of autonomously identifying door handles (a high-touch surface) in unstructured environments and disinfecting them with an ethanol solution. The proposed system is an important step towards harnessing robotics to mitigate the spread of infectious diseases.}
    \label{fig:hero}
\end{figure}
        
    There are several advantages to using UAVs to disinfect surfaces. UAVs that disinfect surfaces could be deployed  en masse, spray high volumes of disinfectant, and limit sanitation workers' exposure to contaminated areas. Given the potential benefits and preliminary deployments, several researchers have conducted investigations into the use of UAVs for spray disinfection - especially since the start of the Covid-19 pandemic. However, the existing literature fails to investigate using UAVs to a) spray disinfectant indoors, b) spray disinfectant with a high level of autonomy, and c) target specific high-touch surfaces with disinfectant. These are important areas of research because, respectively, many pathogens last longer on indoor surfaces sheltered from the elements, unlocking the benefits of mass deployment depends on autonomous functionality, and targeting high-touch surfaces increases the efficiency of sprayed disinfectant. In order to address these shortcomings, this work makes the following important contributions:

\begin{enumerate}
    \item 
    we design, to the best of our knowledge, the first small scale and fully autonomous UAV with disinfectant spraying capabilities,
    \item
    we verify, on a representative sample, that the sprayed disinfectant achieves the coverage necessary to neutralize many common viruses.
\end{enumerate}

    
    

\section{Related work}
    Here we provide a brief overview of existing work related to the use of autonomous UAVs for disinfection. 
    
    
    \subsection{Semi-Autonomous Disinfectant Systems}
        In the wake of the pandemic, there have been several limited investigations into semi-autonomous disinfectant robots and UAVs, and their hardware capabilities. \cite{iot-robot, all-terrain-robot} present tele-operated ground robots equipped with sprayers. \cite{quadcopter-drone, hexcopter-drone, effectiveness-drone} present UAVs equipped with sprayers. Specifically, \cite{effectiveness-drone} and  \cite{quadcopter-drone} look at the trade off between drone weight and volume of disinfectant carried. In both instances, there is mention of indoor flight, but no evidence that the UAVs flew indoors or could do so without direct manual control is provided. Moreover, none of those UAVs have autonomous capabilities greater than the ability to follow preset waypoints, and the effectiveness of the spraying method is not investigated.
    
    \subsection{Spraying Parameters}
        \cite{spray-theory} and \cite{operational-study} investigate how spray parameters including flying height, flying speed, and spray rate impact a disinfectant-spraying UAV's ability to cover a surface. \cite{spray-theory} takes a purely theoretical approach and calculates that varying these parameters can lead to disinfectant coverage between 30 and 0.24 g/m$^2$. \cite{spray-theory} also explores trade-offs between area covered and flight time. \cite{operational-study} investigates how height and speed impact the coverage of a surface with an actual UAV. These lines of research are relevant to our work, because our approach also requires specific spraying parameters to adequately cover a target surface. However, unlike previous work, we propose spraying a relatively smaller target (as opposed to a large swath of ground), so the corresponding spraying parameters are different.
    
    \subsection{Spraying UAVs}
        Finally, most relevant to this work is existing literature on complete UAV spray systems. First, \cite{paint-copter} presents PaintCopter, a fully autonomous UAV that spray paints patterns on surfaces. Notably, this UAV remains tethered during operation and therefore does not address some of the design considerations relevant to our problem. Second, \cite{ozone-drone} presents a UAV equipped with ozone disinfectant for use in outdoor settings. This research is unique because it analyzes the decision to use ozone as a disinfectant and shows that ozone has sterilizing properties against SARS-CoV-2. Similar to this work, we also present an analysis of our disinfectant choice and its sterilizing properties; however, we look at viruses beyond SARS-CoV-2 and use a more widely available disinfectant.

Although quickly growing, the existing body of works on the utilization of autonomous UAVs for the disinfection of indoor surfaces remains incomplete. Most notably, all of the systems mentioned above lack at least one crucial component that would prevent their widespread adoption. Specifically, no existing spraying UAV has the ability to operate autonomously, apply disinfectant in a targeted manner, and safely maneuver indoors. Our work is unique in its attempt to simultaneously address all of these shortcomings, and represents a significant step forward towards the more widespread adoption of UAV systems in addressing concerns related to public health.

\section{UAV Hardware}
	In this section, we introduce the UAV and its hardware. The design was guided by the competing requirements to have full on-board autonomy and spraying capabilities while being small enough to fly indoors. We used a combination of commercially available, custom, and 3D printed parts. To our knowledge, this is the smallest untethered UAV ever built, either commercially or academically, with both autonomous and spraying capabilities, which allow it to operate in unstructured confined indoor settings.
    \subsection{UAV Base}
       The UAV is built on a Holybro S500 quadrotor base. This base is small enough to fly indoors but large enough to mount the compute, sensors, spraying system, and other components. We use a 4-cell battery and 900kV 2812 motors to generate sufficient thrust. At full load, this power system allows the UAV to stay airborne for over 10 minutes. 
	\subsection{Compute}
        The UAV's compute is provided by a Nvidia Jetson Xavier NX. This lightweight edge computing device runs Ubuntu and has a 21 TOPS GPU, which is essential for running object detection on board in real time. The UAV also has a Pixhawk 4 flight controller for lower level motor control and radio communication.
    \subsection{Sensors}
        The UAV uses two visual sensors. The first is the Intel RealSense Depth Camera D435i. This camera provides 30fps $480\times640$ RGBD images. The D channel encodes the distance each pixel is from the camera. The second is the Intel RealSense Tracking Camera T265. This camera provides highly accurate real time pose estimation through use of internal visual odometry. Together, these cameras report the information needed to estimate the UAV’s position and generate a 3D map of its environment. These cameras are mounted to the UAV with a custom 3D printed mount. This mount keeps the cameras at a known transformation relative to each other. Finally, there is a suite of integrated sensors including an IMU in the flight controller.
    \subsection{Spraying}
	The UAV is equipped with a spray system consisting of a tank, pump, and nozzle. The pump (Adafruit, Peristaltic Liquid Pump) draws disinfectant from a 250 mL fuel tank re-purposed to hold the disinfectant. The disinfectant travels through silicone tubing with 2 mm inner diameter and 4 mm outer diameter to a spray nozzle of diameter 80 \textmu m (Victory Innovations, VP50). The whole system is mounted on the UAV with a custom 3D printed mount. The flow rate of water from the nozzle was measured to be 2.6 oz/min.

\section{Spraying Parameters for Surface Disinfection}

There are many parameters that could change how the UAV sprays disinfectant on a target surface. For example, when detecting a door handle to disinfect, the UAV must pick a location relative to the handle from which to spray disinfectant, and that relative location could be modified. In this section we find the spraying parameters needed to neutralize virus particles on a surface with the UAV.

\subsection{Disinfecting a Surface}

To effectively disinfect a surface, we aim to select a disinfectant solution that targets a wide variety of pathogens and is able to inactivate them within a short period of time. The World Health Organization recommends 70\% ethanol as a disinfectant as it inactivates a wide variety of pathogens \cite{ethanol-WHO}. Feline calicivirus (a norovirus surrogate), adenovirus, H1N1, and many coronaviruses are effectively inactivated by 70\% ethanol within 1 minute \cite{ethanol-norovirus,  ethanol-influenza, ethanol-coronaviruses, ethanol-hcov, ethanol-adenovirus-sattar} in carrier tests. Carrier tests determine the effectiveness of a disinfectant on a virus that has been dried on a surface and are more robust than other tests (e.g. suspension tests) in determining the effectiveness of a disinfectant \cite{carrier-test}. For these reasons, we selected 70\% ethanol as the chemical disinfectant solution for the system.

\subsection{Parameter Selection}

   For a spraying-capable UAV to disinfect a surface, it must generate a film of disinfectant on the target surface that persists long enough to inactivate the target pathogens. As discussed in the previous section, this means the UAV, using a 70\% ethanol solution, must generate a film that persists for at least 60 seconds. There are several parameters on a spraying-capable UAV one could use to satisfy this constraint; however, we chose to focus only on spraying distance and spraying time. We chose these parameters because:
   \begin{enumerate}
       \item 
   spraying-capable UAVs can readily adjust their spraying distance and time,
   \item using these parameters does not assume any control over the spraying mechanism,
   \item generating a film that persists for 60 seconds should always be possible by spraying closer to the surface for a longer duration.
  
   \end{enumerate}

\subsection{Spraying Distance}

	We considered two objectives, which compete with each other, when choosing the spraying distance. The first objective is to avoid collisions with obstacles, and the second is for a significant proportion of the droplets to reach the surface. Spray too close and the UAV will collide with walls and handles; spray too far away and the droplets (when sprayed horizontally) will miss the target due to gravity and downdraft caused by the propellers.

\begin{figure} [h]
  \includegraphics[width=\columnwidth, trim=1cm 1cm 1cm 5cm, clip=true]{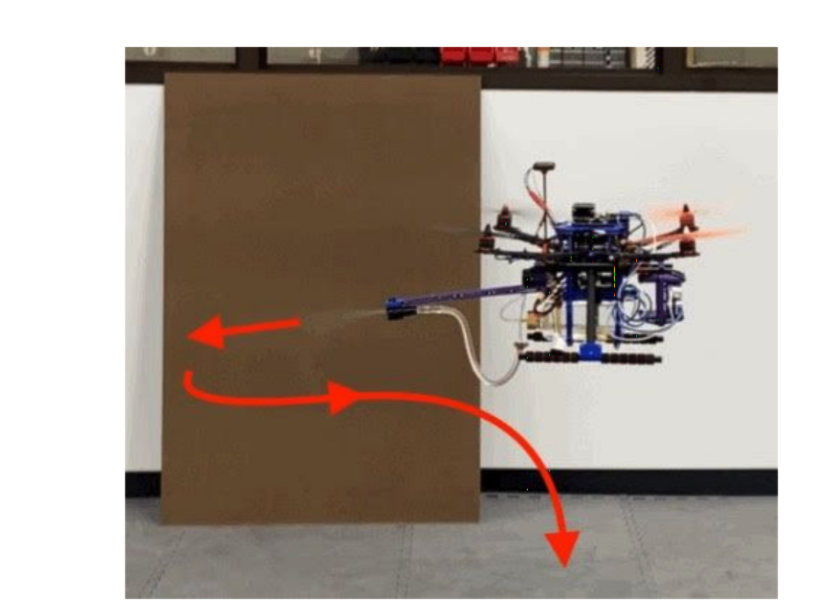}
    \caption{Droplets are swept away by the propellers' downdraft. Spraying too far from a target surface will result in fewer droplets reaching the surface, but spraying too close will result in a greater chance of collision between the UAV and the surface.}
\end{figure}

To quantify the effect of gravity and downdraft, we sprayed a known amount of disinfectant onto a vertical surface attached to a scale while the propellers were running. 

With 30 cm between the spraying nozzle and target surface, 73\% of the droplets (by mass) landed on the surface. Past 30 cm, there was a significant decline in mass of droplets reaching the surface. Therefore, we determined 30 cm was the farthest distance from a surface we could spray while still obtaining a reasonable spray coverage

\subsection{Spraying Time}

\begin{figure} [h]
  \includegraphics[width=\columnwidth, trim=0cm 0cm 0cm 0cm, clip=true]{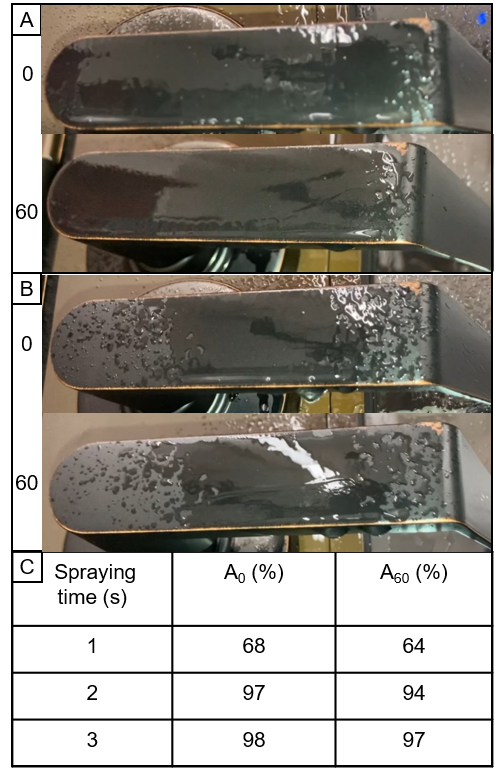}
    \caption{(A) Door handle imaged 0 s and 60 s after spraying for 2 s. (B) Door handle imaged 0 s and 60 s after spraying for 1 s. (C) Percentage of door handle area covered by disinfectant solution 0 s (A\textsubscript{0}) and 60 s (A\textsubscript{60}) after spraying for 1, 2, or 3 seconds. 
}
\label{fig:handle_coverage}
\end{figure}

    We want the UAV to spray the minimal amount of time such that the 70\% ethanol will maintain a film on the target surface for at least one minute to inactivate many common viruses. Since we are spraying from a UAV onto a vertical surface, we expect the ethanol coverage to reduce rapidly due to the combined effects of evaporation and gravity. In order to determine the minimal time required, we sprayed disinfectant onto a vertical surface (the door handle) for 1, 2, and 3 seconds and observed how long the film of ethanol remained on at least 90\% of the exposed door handle surface. We conducted this test with the UAV immobilized (its landing gear was mounted to a chair) and its sprayer 30 cm from the door handle. Before spraying, we turned the UAV's propellers on to create the downdraft that pulls droplets away. We recorded a video of the ethanol solution as it evaporated off the door handle and quantified the coverage area in Figure \ref{fig:handle_coverage} by manually tracing the portion of the area covered by ethanol in ImageJ immediately after spraying and one minute after spraying.

Based on these results (Figure \ref{fig:handle_coverage}), we find that spraying for more than 1 second is required in order to create adequate ($\geq$90\% for 1 min) coverage of the door handle, and spraying for more than 2 seconds uses more disinfectant than necessary. Therefore, we chose 2 seconds for the UAV's spray duration.

\section{Door handle localization and spraying pose}
       To autonomously spray door handles, the UAV must be able to detect, localize, and compute a spraying pose for previously unseen handles in real time. Our approach is inspired from \cite{Arduengo} and uses the same YOLO \cite{Redmon_2016_CVPR} object detection model to provide bounding boxes of handles and doors. When a handle and door are detected, we generate a point cloud representation of each using their respective bonding boxes and the depth channel from the RGBD image. However, unlike \cite{Arduengo}, our perception system has to work at a greater distance from doors to avoid collisions. To do so, we devised a novel method for localizing the handle and computing a spraying pose.
       
        \subsection{Handle Localization}\label{subsec:handleloc}

            The UAV detects handles from a relatively large distance to both reduce collisions and to keep most of the door associated with a handle in the camera's field of view. The downside of detecting from a large distance is that the point cloud representation of a handle region has a low point density. With a low point density and noisy depth channel, it becomes difficult to differentiate points belonging to the actual handle from points belonging to the nearby door surface or door frame. We can approximately localize a handle by computing the raw centroid of the handle region. However, this method results in a large variance in the direction perpendicular to the door due to the noise of the sensor, the ratio of actual handle points and door frame points captured in the handle region, and the potential to include some points from the door frame which may be recessed or protruding (see the red oval in Figure \ref{fig:spraydistance}).
            
                    To address this, we fit a planar model to the door region using the RANSAC algorithm as implemented in the Point Cloud Library \cite{Rusu_ICRA2011_PCL}. Then we project the raw centroid of the handle region to the door plane to reduce the variance in the direction perpendicular to the door. Finally, to calculate our best estimate of a handle's position we use a constant offset in the direction of the door's normal vector to account for a handle's protrusion from the door's surface. The whole process is summarized in Figure \ref{fig:spraydistance}.

        \subsection{Spraying Pose}
\begin{figure} [h]
          \includegraphics[width=\columnwidth, trim=0cm 0cm 0cm 0cm, clip=true]{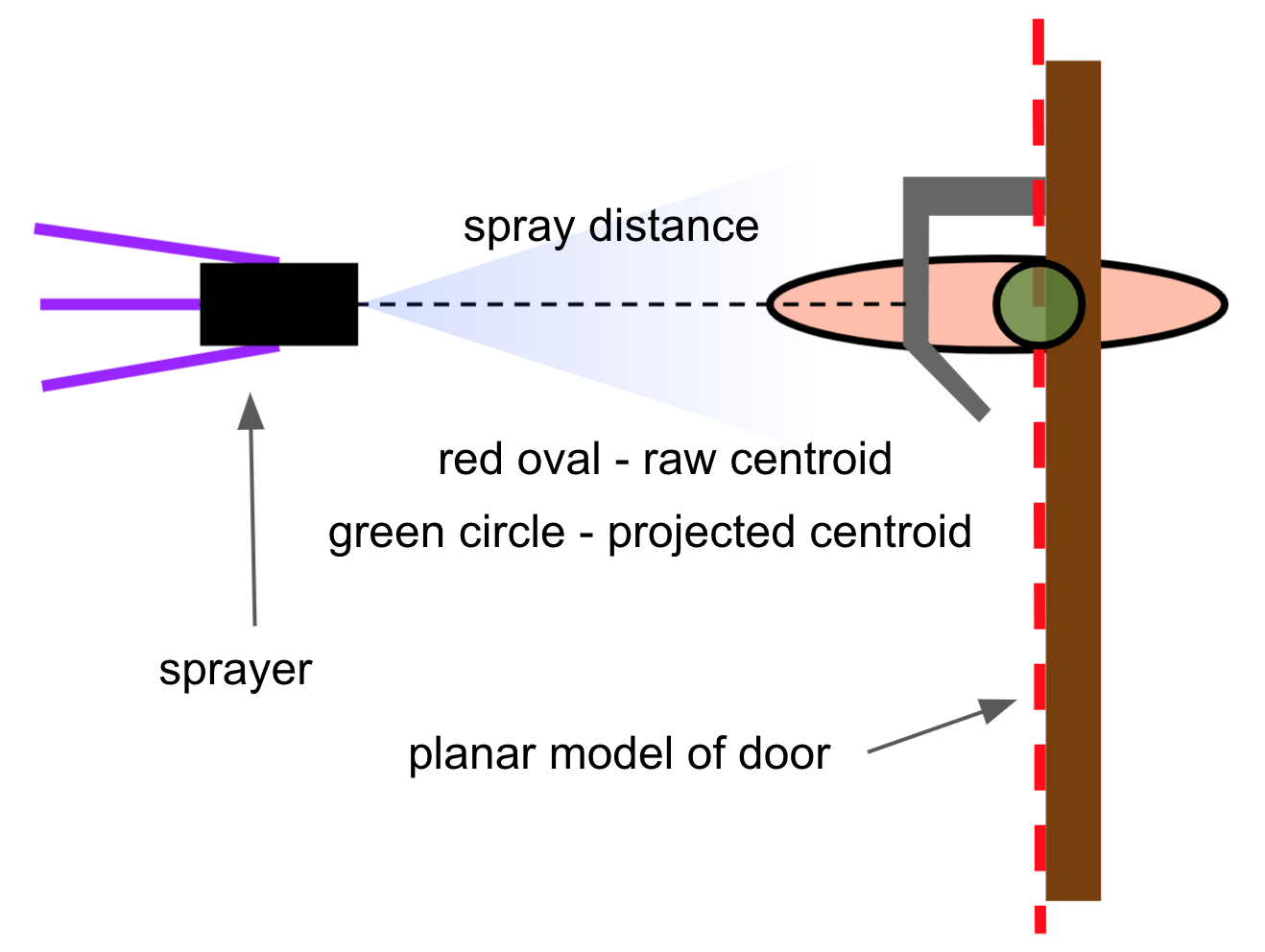}
          \caption{The raw centroid, located somewhere within the red oval, is computed from the door handle point cloud. It is then projected onto the planar model of the door. The spray pose is computed so the sprayer is facing the handle and 30 cm from the handle.}
          \label{fig:spraydistance}
        \end{figure}

            We compute the spraying position from the handle position and the door normal such that the when in the spraying position, the UAV's spraying nozzle points perpendicular to the door and is $30$cm from the handle. See Figure \ref{fig:spraydistance} for a visualization of the spraying pose relative to a handle and door.

    \section{Planning strategy}
        To autonomously spray door handles, the UAV must have several advanced software features including mapping, collision free path planning, path following, and a state machine. In this section we provide an overview of these software components. 
        
        \subsection{Mapping}
            The UAV creates and updates a 3D map of its environment in the form of an voxel grid, a efficient volumetric representation of free, empty, and unexplored space. This map is created from point cloud measurements paired with the UAV's pose during measurement. Figure \ref{fig:map_path} shows a map of a hallway that the UAV generated.

        \subsection{Path Planning}
\begin{figure} [h]
          \includegraphics[width=\columnwidth, trim=0cm 0cm 0cm 0cm, clip=true]{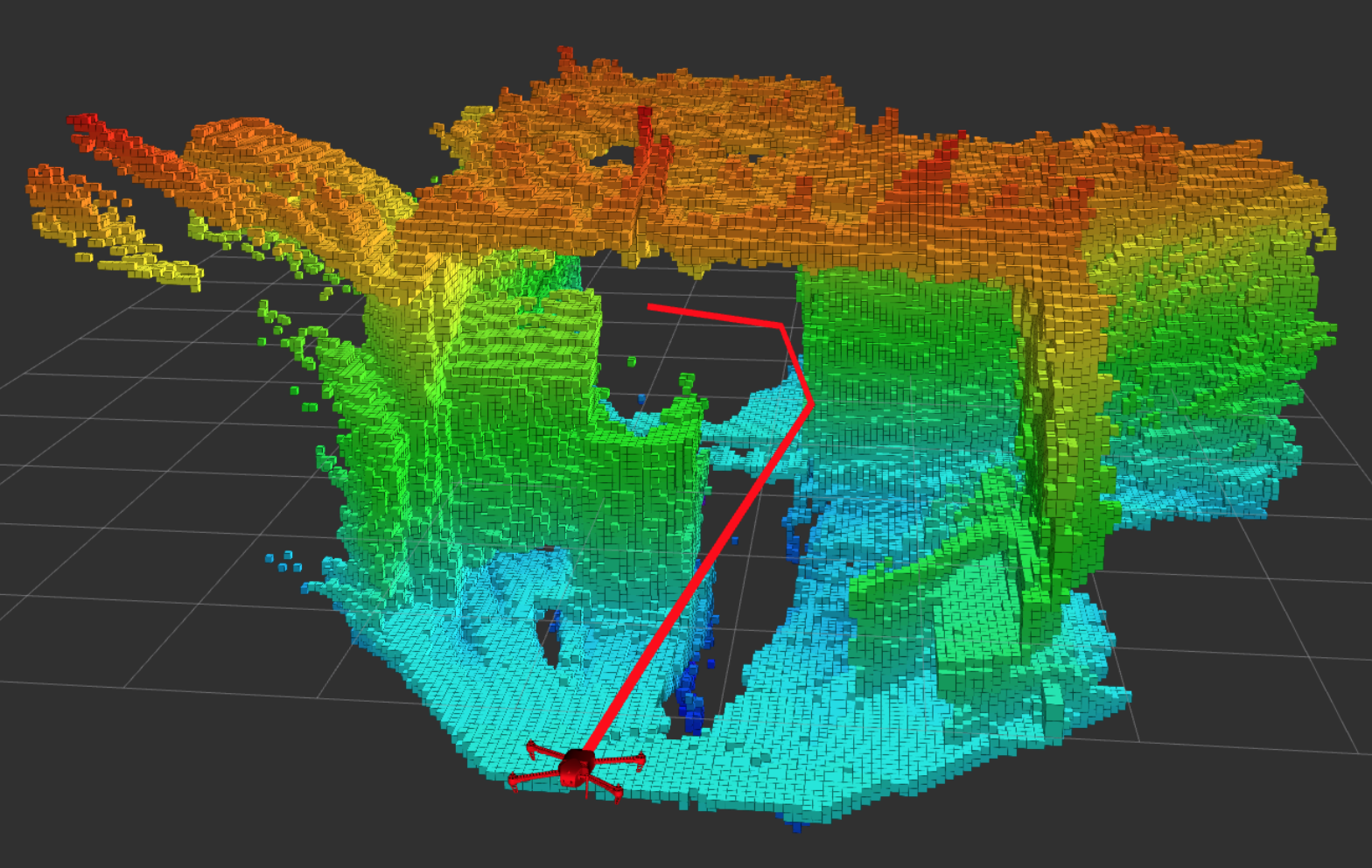}
            \caption{A visual representation of the 3D occupancy map estimated using the UAV's depth and tracking cameras. Also pictured is a path computed using RRT*  that avoids the the obstacles in the map.}
            \label{fig:map_path}
        \end{figure}

            Given a start position, end position, and 3D map of the environment, we must plan a 3D path from start to end that will avoid all collisions. To do so, we use the OMPL \cite {sucan2012the-open-motion-planning-library} implementation of RRT* \cite{KaramanFrazzoli2011}, an efficient search algorithm in complex spaces. This algorithm requires the UAV to predict if certain positions will result in a collision, which it does by modeling itself as a rectangular cuboid and checking for collisions with the voxel grid. Figure \ref{fig:map_path} shows a path the UAV planned around an obstacle.
            
        \subsection{Path Following}
            Given a series of waypoints from the path planning module, the UAV sends these waypoints to the PX4 flight controller, which then follows them using a cascaded PID-based controller.
        
        \subsection{State Machine}
\begin{figure} [h]
          \includegraphics[width=\columnwidth, trim=0cm 0cm 0cm 0cm, clip=true]{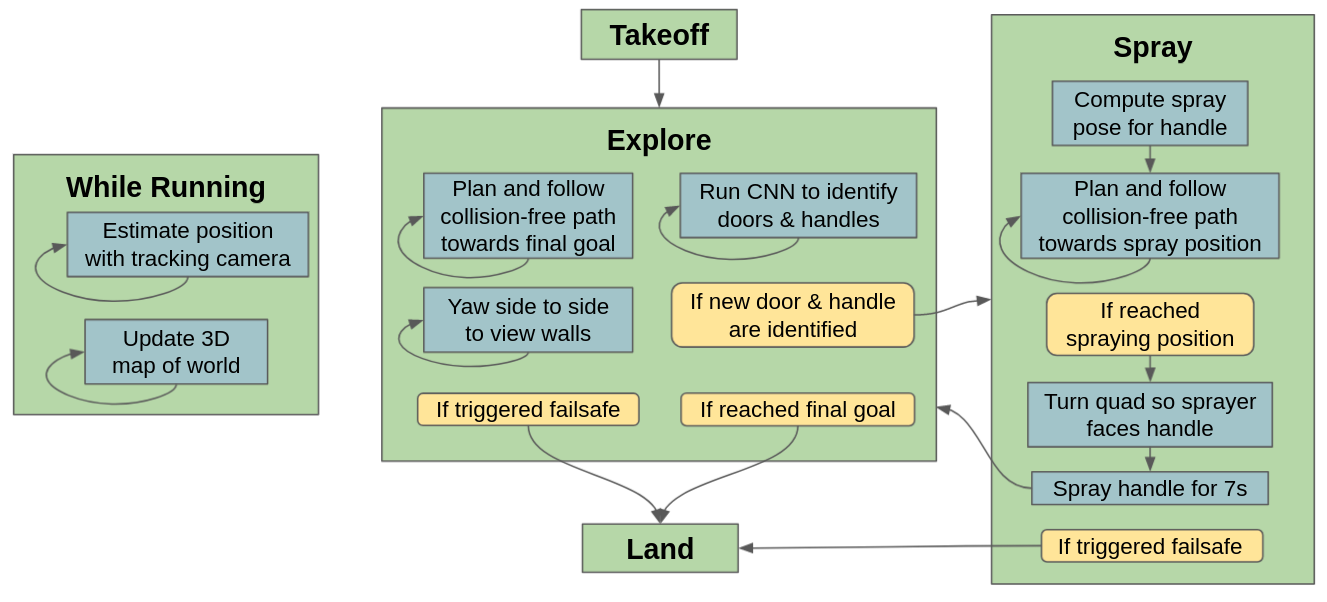}
            \caption{At all times, the UAV is estimating its position and building a map of the world. To spray door handles, the UAV first takes off and then transitions between searching for door handles in explore mode and spraying the handles in spray mode. In explore mode, the UAV travels towards a final goal while yawing to see handles on walls. In spray mode, the UAV computes a spraying pose for a detected handle and disinfects the handle. The UAV lands once it reaches its final goal or encounters a safety concern.}
            \label{fig:state_machine}
        \end{figure}

            At a high level, the UAV can takeoff, explore towards a set final goal, detour to spray door handles as it sees them, and land when it reaches the final goal (follow along in Figure \ref{fig:state_machine}). In all stages of flight, the UAV continuously estimates its position using the tracking camera and builds a 3D map of the world using its position and the depth camera measurements. The UAV starts by taking off with a preset final goal. Once takeoff is complete, the UAV immediately enters explore mode. In explore mode, the UAV plans and follows a collision free path towards the final goal. The UAV assumes that following this path will lead it through a hallway that could contain handles on either side. While following this path, the UAV yaws side to side in order to scan both walls. When the UAV detects both a door and a handle on the door with its CNN, the UAV stops exploring and switches to spray mode. In spray mode, the UAV computes the spray pose as described in section 5.2 and travels towards this pose while keeping the handle in its field of view. Once the UAV is within one meter of the spray position, it rotates to aim the sprayer towards the handle, waits to be within a small threshold of the desired spray position, and then turns the sprayer on for 2 seconds. When spraying is finished, the UAV returns to the center of the hallway and continues in explore mode. If the UAV triggers a fail safe like low battery or inability to reach the final goal, it will land. If the UAV does reach the final goal, it will land having successfully completed its mission.

        Our code and a video demonstration of the UAV are available to view at github.com/StanfordASL/surface-disinfect.

\section{Experimental validation}
    Given the context in which our proposed system is meant to be deployed, it is critical that we validate its ability to consistently and accurately localize and spray chosen surfaces in its environment. Indeed, substituting manual disinfection for an automated solution similar to the one presented in this work could exacerbate the spread of harmful pathogens if the autonomous system chosen were unable to achieve a minimum level of reliability. To this effect, we tested our proposed system in a motion capture environment that contained a door and a door handle. The motion capture system allowed us to validate the performance of the autonomous system via ground truth information.
        
\begin{figure} [h]
          \includegraphics[width=\columnwidth, trim=0cm 0cm 1cm 4cm, clip=true]{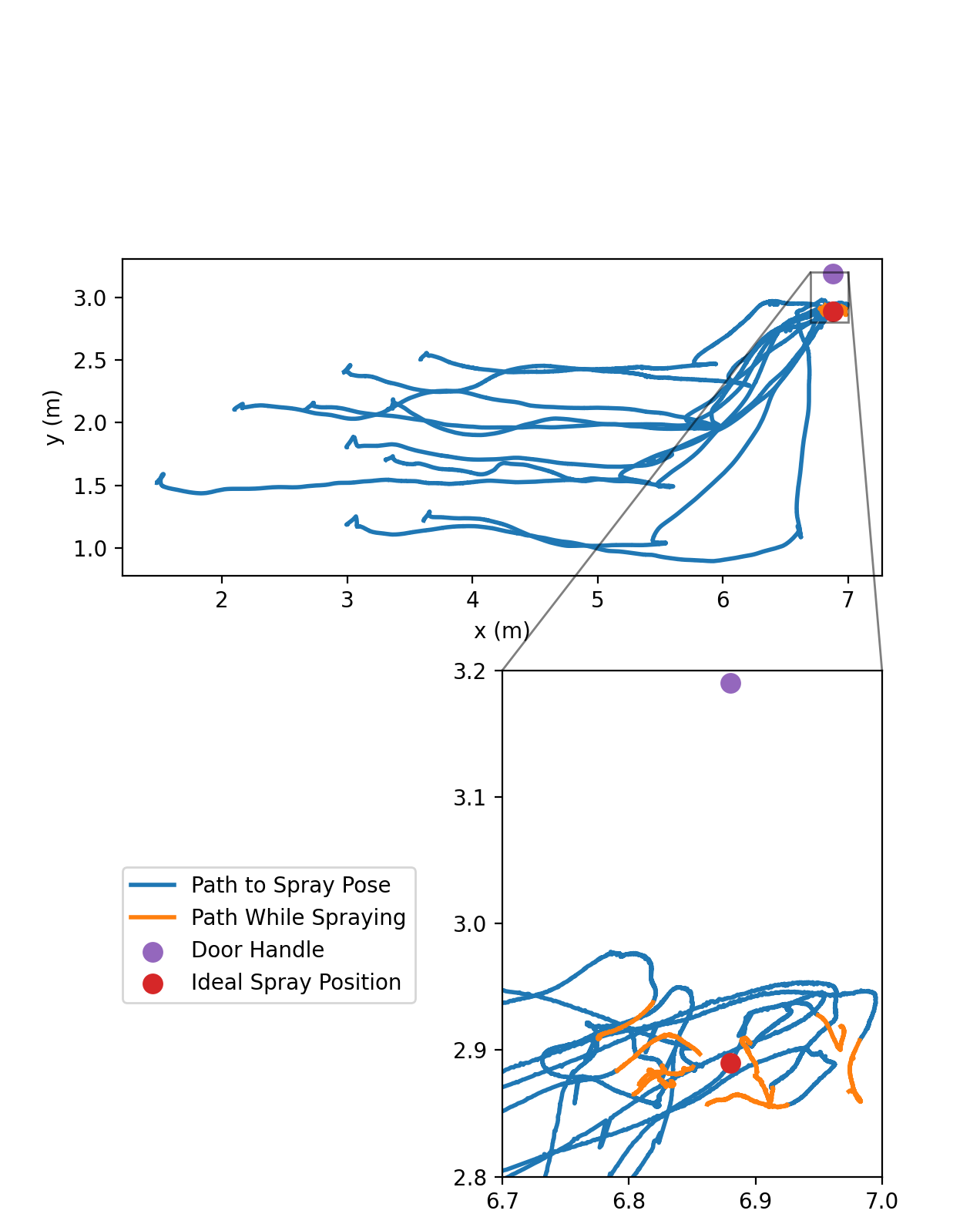}
            \caption{The Drone's path during ten consecutive trials. The abrupt left turn in the blue paths show when the drone starts tracking towards the computed spray pose. The orange segments show the drone's path during the 2 second spray time.}
            \label{fig:val}
        \end{figure}

    First, we determined where the target spray pose should be by pointing the UAV's sprayer directly at the door handle at a distance of 30 cm and recording the position reported by the motion capture system. We then ran 10 trials where the UAV started at different positions and headed towards a final goal that would lead it past the door and door handle. In all 10 trials, the UAV autonomously found and sprayed the door handle. See Figure \ref{fig:val} for a top down view of each path the UAV took towards the handle. 

\begin{figure} [h]
          \includegraphics[width=\columnwidth, trim=0cm 0cm 0cm 0cm, clip=true]{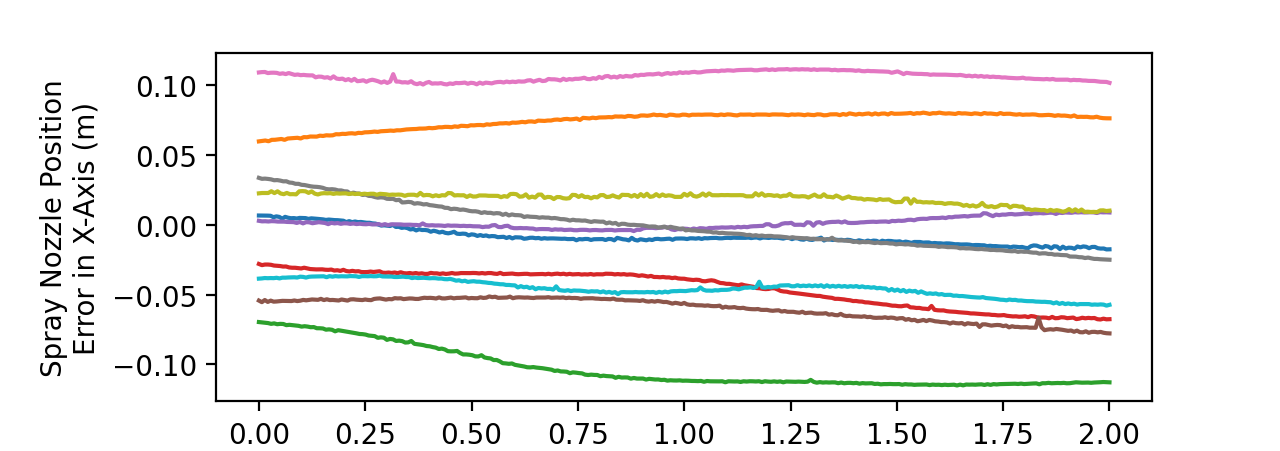}
        \includegraphics[width=\columnwidth, trim=0cm 0cm 0cm 0cm, clip=true]{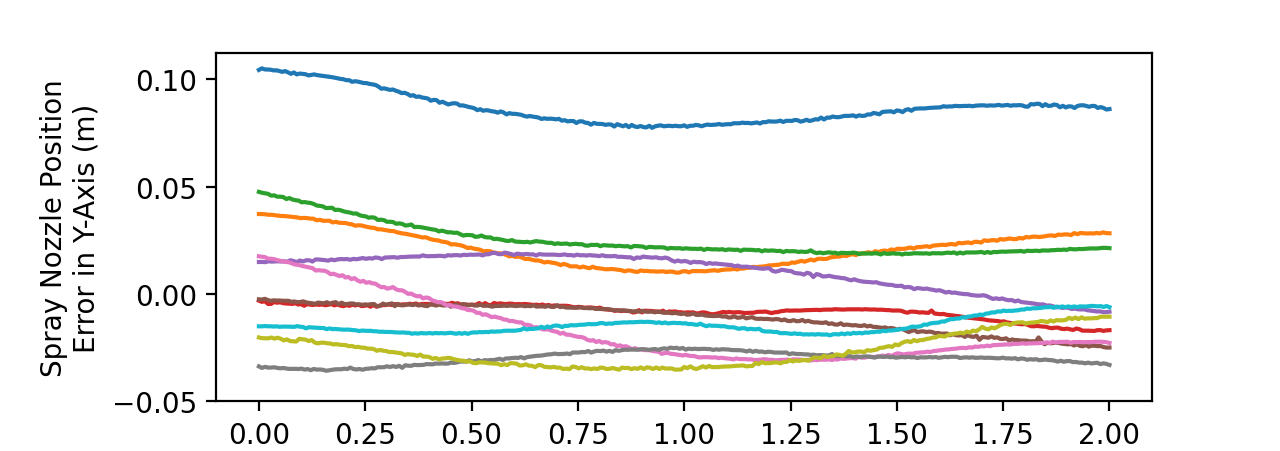}
            \caption{Plotted above is the difference between the ideal spray position for the UAV's spray nozzle and the actual position during the 2 seconds of spraying. This indicates that overall performance could be most improved by enhancing target localization as opposed to position control.}
            \label{fig:pos_errors}
        \end{figure}

    While the drone was spraying the handle, its average distance from the target spray pose was $6.8$ cm. In Figure \ref{fig:pos_errors} we plot the difference between the spray nozzle's ideal position and its actual position during each run. Notice that the deviation in the nozzle's position during an individual trial is relatively small compared to the deviation between trials. This indicates that, relatively speaking, the drone is better at holding a position than it is at localizing door handles in its environment. This further suggests that overall performance improvement could be most easily attained by developing better localization strategies than the one presented in section \ref{subsec:handleloc}.

\section{Conclusion}

    In this work, we present a UAV that can autonomously find door handles and spray them with disinfectant in unstructured, novel indoor environments. We showed that this UAV can reliably find and target door handles with disinfectant spray. We also demonstrated that disinfectant spray from this UAV could neutralize many common viruses on the surface of a door handle without using excessive disinfectant.
        
    As this UAV system was novel in many regards, our work exposes several avenues for future work. First, we found that building a system that was small enough to operate in indoor environments (such as hallways) put significant limitations on the UAV's ability to carry large amounts of disinfectant. Indeed, in its current configuration, the UAV is carrying enough disinfectant to spray only about 20 door handles. Additionally, future iterations of the UAV would ideally be able to spray a larger variety of surfaces. Moreover, the flight time of the UAV is dramatically affected by the payload requirements of this task. One potentially promising solution would be to use a UAV tethered to a mobile ground robot such that the UAV's compute, power, and disinfectant could all be stored on the ground. This would allow for vastly increased disinfectant capacity and flight time, while maintaining the maneuverability and targeting benefits of a UAV. Finally, future iterations of this UAV could allow for full coverage of all surfaces on a target object, including any crevices or surfaces that are difficult to reach, by having an adjustable nozzle that allows the UAV to spray the object or surface from different directions. 

    Although our proposed system suffers from some limitations, it helps lay the groundwork to develop future fleets of UAVs that reduce the spread of pathogens via fomites, and for a broader adoption of autonomous systems to tackle key societal challenges related to public health.

\section*{ACKNOWLEDGEMENT}

We thank the Stanford UAV club for their thoughtful advice on the propulsion system. This work was supported by a National Science Foundation grant CBET-2030390.

\printbibliography

@Preamble{"\newcommand{\noopsort}[1]{} " #
"\newcommand{\printfirst}[2]{#1} " #
"\newcommand{\singleletter}[1]{#1} " #
"\newcommand{\switchargs}[2]{#2#1} "}

@String { jrn_SAGE_IJRR             = {{Int.\ Journal of Robotics Research}} }

@Article{KaramanFrazzoli2011,
  Title                    = {Sampling-based Algorithms for Optimal Motion Planning},
  Author                   = {Karaman, S. and Frazzoli, E.},
  Journal                  = jrn_SAGE_IJRR,
  Year                     = {2011},

  Number                   = {7},
  Pages                    = {846--894},
  Volume                   = {30}
}

@InProceedings{Redmon_2016_CVPR,
author = {Redmon, Joseph and Divvala, Santosh and Girshick, Ross and Farhadi, Ali},
title = {You Only Look Once: Unified, Real-Time Object Detection},
booktitle = {Proceedings of the IEEE Conference on Computer Vision and Pattern Recognition (CVPR)},
month = {June},
year = {2016}
}

@unpublished{Arduengo,
  Title                    = {Robust and Adaptive Door Operation with a Mobile Robot},
  Author                   = {Miguel Arduengo and Carme Torras and Luis Sentis},
  howpublished                 = {arXiv preprint arXiv:1902.09051},
  Year                     = {2019}
}

@article{sucan2012the-open-motion-planning-library,
    Author = {Ioan A. Sucan and Mark Moll and Lydia E. Kavraki},
    Doi = {10.1109/MRA.2012.2205651},
    Journal = {{IEEE} Robotics \& Automation Magazine},
    Month = {December},
    Number = {4},
    Pages = {72--82},
    Title = {The {O}pen {M}otion {P}lanning {L}ibrary},
    Note = {https://ompl.kavrakilab.org},
    Volume = {19},
    Year = {2012}
}

@report{cdc-surface-transmission,
    Author = {Center For Disease Control and Prevention},
    Title = {Science Brief: SARS-CoV-2 and Surface (Fomite) Transmission for Indoor Community Environments},
    Month = {April},
    Year = {2021}
}

@online{nigeria-disenfect,
    Title = {COVID-19: Nigerian Startup Beat Drone deploys drones to disinfect communities},
    url = {https://www.africanreview.com/manufacturing/industry/covid-19-nigerian-startup-beat-drone-deploys-drones-to-disinfect-communities},
    urldate = {2021-05-25}
}

@online{china-disenfect,
    Title = {3 ways china is using drones to fight coronavirus},
    url = {https://www.weforum.org/agenda/2020/03/three-ways-china-is-using-drones-to-fight-coronavirus},
    urldate = {2021-05-25}
}

@online{usa-disenfect,
    Title = {EagleHawk Deploys Disinfectant Drones to Sanitize Facilities},
    url = {https://dronelife.com/2020/05/05/disinfectant-drones-eaglehawk},
    urldate = {2021-05-25}
}

@online{india-disenfect,
    Title = {India uses drones to disinfect virus hotspot as cases surge},
    Note = {https://medicalxpress.com/news/2020-05-drones-disinfect-indian-pandemic-hotspot},
    urldate = {2021-05-25}
}

@online{spain-disenfect,
    Title = {Spain’s military uses DJI agricultural drones to spray disinfectant in fight against Covid-19},
    url = {https://www.scmp.com/tech/gear/article/3077945/spains-military-uses-dji-agricultural-drones-spray-disinfectant-fight},
    urldate = {2021-05-25}
}

@online{france-disenfect,
Title = {First use of drones to fight coronavirus in France},
url = {https://www.youtube.com/watch?v=scgmi2loSWg},
urldate = {2021-05-25}
}

@online{indonesia-disenfect,
Title = {Coronavirus: Indonesia uses drones to disinfect dense areas},
url = {https://www.youtube.com/watch?v=IjhN3hzgLLg},
urldate = {2021-05-25}
}

@InProceedings{ozone-drone,
    Title = {Assessing the potential of unmanned aerial vehicle spraying of aqueous ozone as an outdoor disinfectant for SARS-CoV-2},
    Author = {Simon Albert and Alberto Amarilla and Ben Trollope, Julian D.J. Sng and Yin Xiang Setoh and Nathaniel Deering, Naphak Modhiran and Sung-Hsia Weng and Maria C. Melo and Nicholas Hutley and Avik Nandy and Michael J. Furlong and Paul R. Young and Daniel Watterson and Alistair R. Grinham and Alexander A. Khromykh},
    booktitle = {Enviromental Research},
    Month = {February},
    Year = {2021}
}

@InProceedings{all-terrain-robot,
    Title = {All-terrain mobile robot disinfectant sprayer to decrease the spread of COVID-19 in open area},
    Author = {Prisma Megantoro and Herlambang Setiadi and Brahmantya Aji Pramudita},
    booktitle = {International Journal of Electrical and Computer Engineering},
    Month = {June},
    Year = {2021}
}

@InProceedings{iot-robot,
    Title = {Design and Development of Spray Disinfection System to Combat Coronavirus (Covid-19) Using IoT Based Robotics Technology},
    Author = {M. N. Mohammed and Elina abd aziz and Inam Sameh Arif and S. Al-Zubaidi and Siti Humairah and Kamarul Bahrain and Sairah A.K and Eddy Yusuf},
    booktitle = {Revista Argentina de Clínica Psicológica},
    Year = {2020}
}

@InProceedings{quadcopter-drone,
    Title = {Disinfectant Spraying System with Quadcopter Type Unmanned Aerial Vehicle Technology as an Effort to Break the Chain of the COVID-19 Virus},
    Author = {Dwi Mutiara Harfina and Zaini Zaini and Wisnu Joko Wulung},
    booktitle = {Journal of Robotics and Control},
    Month = {November},
    Year = {2021}
}

@InProceedings{paint-copter,
    Title = {PaintCopter: An Autonomous UAV for Spray Painting on Three-Dimensional Surfaces},
    Author = {Anurag Sai Vempati and Mina Kamel and Nikola Stilinovic and Qixuan Zhang and Dorothea Reusser and Inkyu Sa and Juan Nieto and Roland Siegwart and Paul Beardsley},
    booktitle = {IEEE Robotics and Automation Letters},
    Month = {June},
    Year = {2018}
}

@InProceedings{effectiveness-drone,
    Title = {The effectiveness of disinfectant spraying based on drone technology},
    Author = {T Andrasto et al},
    booktitle = {IOP Conference Series: Earth and Enviromental Science},
    Year = {2021}
}

@InProceedings{hexcopter-drone,
    Title={Sanitization using Hexacopter Autonomous Drone},
    Author = {K Ramesh and B Priya Dharshini and K Haridass and S Deepak Kumar and R Gokul Raj and V Hariprasad},
    booktitle = {IOP Conference Series: Materials Science and Engineering},
    Year = {2021}
}

@InProceedings{operational-study,
    Title = {Operational Study of Drone Spraying Application for the Disinfection of Surfaces against the COVID-19 Pandemic},
    Author = {Higinio González Jorge and Luis Miguel González de Santos and Noelia Fariñas Álvarez and Joaquin Martínez Sánchez and Fermin Navarro Medina},
    booktitle = {MDPI Drones},
    Month = {March},
    Year = {2021}
}

@online{spray-theory,
  Title                    = {Drone Application for Spraying Disinfection Liquid Fighting against Covid-19 Pandemic – Theoretic Approach},
  Author                   = {Ágoston Restás and István Szalkai and Gyula Óvári},
  Month                     = {March},
  Year                     = {2021},
}

@online{norovirus,
    Title = {Norovirus},
    urldate = {2021-05-18},
    Organization = {Centers for Disease Control and Prevention},
    url={https://www.cdc.gov/norovirus/index.html}
}

@online{adenoviruses,
    Title = {Adenoviruses},
    url={https://www.cdc.gov/adenovirus/about/transmission.html},
    Organization = {Centers for Disease Control and Prevention},
    urldate = {2021-05-18}
}

@online{h1n1,
    Title = {H1n1 Flu},
    url={https://www.cdc.gov/h1n1flu/qa.htm},
    Organization = {Centers for Disease Control and Prevention},
    urldate = {2021-05-18}
}

@article{fomite-transmission-norovirus-influenza-rhinovirus,
    Author = {Alicia N.M. Kraay and Michael A.L. Hayashi and Nancy Hernandez-Ceron and Ian H. Spicknall and Marisa C. Eisenberg and Rafael Meza and Joseph N.S. Eisenberg},
    Doi = {10.1186/s12879-018-3425-x},
    Journal = {BMC Infectious Diseases},
    Month = {October},
    Number = {540},
    Title = {Fomite-mediated transmission as a sufficient pathway: a comparative analysis across three viral pathogens},
    Volume = {18},
    Year = {2018}
}

@article{fomite-transmission-adenovirus,
    Author = {Ana Carolina Ganime and Filipe A. Carvalho-Costa and Marisa Santos and Rubens {Costa Filho} and José Paulo G. Leite and Marize P. Miagostovich},
    Title = {Viability of human adenovirus from hospital fomites},
    Doi = {10.1002/jmv.23907},
    Journal = {Journal of Medical Virology},
    Month = {December},
    Year = {2014},
    Volume = {86},
    Number = {12},
    Pages = {2065--2069},
}

@incollection{ethanol-WHO,
	author	= "World Health Organization",
	title	= "Use of disinfectants:alcohol and bleach",
	booktitle= "Infection prevention and control of epidemic- and pandemic-prone acute respiratory infections in health care",
	publisher= "World Health Organization",
	chapter= "Annex G",
	pages	= "65",
	year	= "2014",
}

@article{ethanol-coronaviruses,
    Author = {Nicolas Castaño and Seth C. Cordts and Myra {Kurosu Jalil} and Kevin S. Zhang and Saisneha Koppaka and Alison D. Bick and Rajorshi Paul and Sindy K. Y. Tang},
    Title = {Fomite Transmission, Physicochemical Origin of Virus–Surface Interactions, and Disinfection Strategies for Enveloped Viruses with Applications to SARS-CoV-2},
    Doi = {10.1021/acsomega.0c06335},
    Journal = {ACS Omega},
    Month = {March},
    Year = {2021},
    Volume = {6},
    Number = {10},
    Pages = {6509--6527},
}

@article{ethanol-influenza,
    Author = {Eun Kyo Jeong and Jung Eun Bae and In Seop Kim},
    Title = {Inactivation of influenza A virus H1N1 by disinfection process},
    Doi = {10.1016/j.ajic.2010.03.003},
    Journal = {American Journal of Infection Control},
    Month = {June},
    Year = {2010},
    Volume = {38},
    Number = {5},
    Pages = {354--360},
}

@article{ethanol-norovirus,
    Author = {Yashpal S. Malik and Sunil Maherchandani and Sagar M. Goyal},
    Title = {Comparative efficacy of ethanol and isopropanol against feline calicivirus, a norovirus surrogate},
    Doi = {10.1016/j.ajic.2005.05.012},
    Journal = {American Journal of Infection Control},
    Month = {February},
    Year = {2006},
    Volume = {34},
    Number = {1},
    Pages = {31--35},
}

@article{ethanol-hcov,
    Author = {C. Meyers and R. Kass and D. Goldenberg and J. Milici and S. Alam and R. Robison},
    Title = {Ethanol and isopropanol inactivation of human coronavirus on hard surfaces},
    Doi = {10.1016/j.jhin.2020.09.026},
    Journal = {Journal of Hospital Infection},
    Month = {January},
    Year = {2021},
    Volume = {107},
    Pages = {45--49},
}

@article{illness-norovirus-hall,
    Author = {Aron J. Hall and Ben A. Lopman and Daniel C. Payne and Manish M. Patel and Paul A. Gastañaduy and Jan Vinjé and Umesh D. Parashar},
    Title = {Norovirus Disease in the United States},
    Doi = {10.3201/eid1908.130465},
    Journal = {Emerging Infectious Diseases},
    Month = {August},
    Year = {2013},
    Volume = {19},
    Number = {8},
    Pages = {1198--1205},
}

@article{fomite-transmission-norovirus-h1n1-SARS-Lei,
    Author = {H. Lei and Y. Li and S. Xiao and C.-H. Lin and S. L. Norris and D. Wei and Z. Hu and S. Ji},
    Title = {Routes of transmission of influenza A H1N1, SARS CoV, and norovirus in air cabin: Comparative analyses},
    Doi = {10.1111/ina.12445},
    Journal = {Indoor Air},
    Month = {April},
    Year = {2018},
    Volume = {28},
    Number = {3},
    Pages = {394--403},
}

@article{fomite-transmission-norovirus-Repp,
    Author = {Kimberly K. Repp and William E. Keene},
    Title = {A Point-Source Norovirus Outbreak Caused by Exposure to Fomites},
    Doi = {10.1093/infdis/jis250},
    Journal = {The Journal of Infectious Diseases},
    Month = {June},
    Year = {2012},
    Volume = {205},
    Number = {11},
    Pages = {1639-–1641},
}

@article{carrier-test,
    Author = {Syed A Sattar and V.Susan Springthorpe and Olusola Adegbunrin and A.Abu Zafer and Maria Busa},
    Title = {A disc-based quantitative carrier test method to assess the virucidal activity of chemical germicides},
    Doi = {10.1016/S0166-0934(03)00192-7},
    Journal = {Journal of Virological Methods},
    Month = {September},
    Year = {2003},
    Volume = {112},
    Number = {1-2},
    Pages = {3-–12},
}

@article{fomite-transmission-hcov-doorknobs,
    Author = {Tania S. Bonny and Saber Yezli and John A. Lednicky},
    Title = {Isolation and identification of human coronavirus 229E from frequently touched environmental surfaces of a university classroom that is cleaned daily},
    Doi = {10.1016/j.ajic.2017.07.014},
    Journal = {American Journal of Infection Control},
    Month = {January},
    Year = {2018},
    Volume = {46},
    Number = {1},
    Pages = {105-–107},
}

@article{fomite-transmission-MERS,
    Author = {Shui Shan Lee and Ngai Sze Wong},
    Title = {Probable transmission chains of Middle East respiratory syndrome coronavirus and the multiple generations of secondary infection in South Korea},
    Doi = {10.1016/j.ijid.2015.07.014},
    Journal = {International Journal of Infectious Diseases},
    Month = {September},
    Year = {2015},
    Volume = {38},
    Pages = {65-–67},
}

@article{ethanol-adenovirus-sattar,
    Author = {S. A. Sattar and V. S. Springthorpe and Y. Karim and P. Loro},
    Title = {Chemical disinfection of non-porous inanimate surfaces experimentally contaminated with four human pathogenic viruses},
    Doi = {10.1017/S0950268800030211},
    Journal = {Epidemiology \& Infection},
    Month = {June},
    Year = {1989},
    Volume = {102},
    Number = {3},
    Pages = {493--505},
}

@InProceedings{Rusu_ICRA2011_PCL,
  author    = {Radu Bogdan Rusu and Steve Cousins},
  title     = {{3D is here: Point Cloud Library (PCL)}},
  booktitle = {{IEEE International Conference on Robotics and Automation (ICRA)}},
  month     = {May 9-13},
  year      = {2011},
  address   = {Shanghai, China},
  publisher = {IEEE}
}

@article{covid-modeling,
    Author = {Alicia N.M. Kraay and Michael A.L. Hayashi and David M. Berendes and Julia S. Sobolik and Juan S. Leon and Benjamin A. Lopman},
    Title = {Risk of fomite-mediated transmission of SARS-CoV-2 in child daycares, schools, and offices: a modeling study},
    Doi = {10.1101/2020.08.10.20171629},
    Journal = {medRxiv},
    Month = {August},
    Year = {2020},
}
\end{document}